\documentclass{article}

    \usepackage[preprint]{neurips_2023}

\usepackage[utf8]{inputenc} % allow utf-8 input
\usepackage[T1]{fontenc}    % use 8-bit T1 fonts
\usepackage{hyperref}       % hyperlinks
\usepackage{url}            % simple URL typesetting
\usepackage{booktabs}       % professional-quality tables
\usepackage{amsfonts}       % blackboard math symbols
\usepackage{nicefrac}       % compact symbols for 1/2, etc.
\usepackage{microtype}      % microtypography
\usepackage{xcolor}         % colors
\usepackage{graphicx}
\usepackage{tikz}
\usepackage{comment}
\usepackage{amsmath,amssymb} % define this before the line numbering.
\usepackage{color}
\usepackage{algorithm}
\usepackage{algpseudocode}
\usepackage{graphics} % for pdf, bitmapped graphics files
\usepackage{epsfig} % for postscript graphics files
\usepackage{hyperref}
\usepackage{balance}
\usepackage{mathptmx} % assumes new font selection scheme installed
\usepackage{interval}
\usepackage{threeparttable}
\usepackage{booktabs}
\usepackage{adjustbox}
\usepackage{multicol}
\usepackage{lipsum}
\usepackage{graphicx}
\usepackage{booktabs}
\usepackage{caption}
\usepackage{pgfplots}
\usepackage{pgfplotstable}
\usepackage{subcaption}
\usepackage{tikz}

\title{RealCraft: Attention Control as A Tool for Zero-Shot Consistent Video Editing}

\author{%
  Shutong Jin \\
  KTH \\
  \texttt{shutong@kth.se} \\
  \And
  Ruiyu Wang \\
  KTH \\
  \texttt{ruiyuw@kth.se} \\
  \And
  Florian T. Pokorny \\
  KTH \\
  \texttt{fpokorny@kth.se} \\
}

\begin{document}

\maketitle

\begin{abstract}
  Even though large-scale text-to-image generative models show promising performance in synthesizing high-quality images, applying these models directly to image editing remains a significant challenge. This challenge is further amplified in video editing due to the additional dimension of time. This is especially the case for editing real-world videos as it necessitates maintaining a stable structural layout across frames while executing localized edits without disrupting the existing content. In this paper, we propose \textit{RealCraft}, an attention-control-based method for zero-shot real-world video editing. By swapping cross-attention for new feature injection and relaxing spatial-temporal attention of the editing object, we achieve localized shape-wise edit along with enhanced temporal consistency. Our model directly uses Stable Diffusion and operates without the need for additional information. We showcase the proposed zero-shot attention-control-based method across a range of videos, demonstrating shape-wise, time-consistent and parameter-free editing in videos of up to 64 frames.

\end{abstract}

% \begin{figure}
% \centering
% \includegraphics[height=10cm]{figures/first_figure_v3.pdf}
% \caption{\textit{RealCraft} performs precise and consistent shape editing. In example (a), a 40-frame long video is edited, achieving notable backgrounds and detailed changes (e.g. flower) with high temporal consistency. Examples (b) - (d) demonstrate precise shape editing: the car in (b), the ear and tail of the dog in (c) and the boat in (d) are significantly altered according to the editing prompt.}
% \label{fig:first_figure}
% \end{figure}

\section{Introduction}
Recent advancements in large-scale text-driven diffusion models such as DALL·E 3 \cite{betker2023improving}, Imagen \cite{saharia2022photorealistic}, Stable Diffusion \cite{rombach2022high}, and SORA \cite{videoworldsimulators2024} have shown remarkable capabilities in generating high-quality and diverse visual content with only textual inputs. However, there is no smooth transfer from visual generation to text-driven editing, since diffusion models only know what to change but not what to preserve with given text prompts \cite{parmar2023zero}. Additionally, compared with synthetic data, semantic editing of real-world images or videos poses extra challenges in maintaining structural consistency and preserving localization. This is due to the fact that, for synthetic editing, the original images or videos are typically sampled from the latent space of diffusion models, based on source textual prompts \cite{hertz2022prompt}. Semantic information of visual content is therefore well preserved during the sampling process. Real-world images and videos on the other hand feature cluttered scenes, occluded objects or moving camera views \cite{bar2022text2live} whose semantic information may not be fully captured by diffusion models. Furthermore, for video editing, there is a trade-off between maintaining temporal consistency and enforcing localized and significant changes \cite{lee2023shape}. Thus, most editing of real-world videos to date is still limited to style transfer \cite{yang2023rerender}. For simplicity, we refer to real-world videos (as opposed to synthetically generated videos) as real videos in the following paper.

\begin{figure}[!ht]
\centering
\includegraphics[height=10cm]{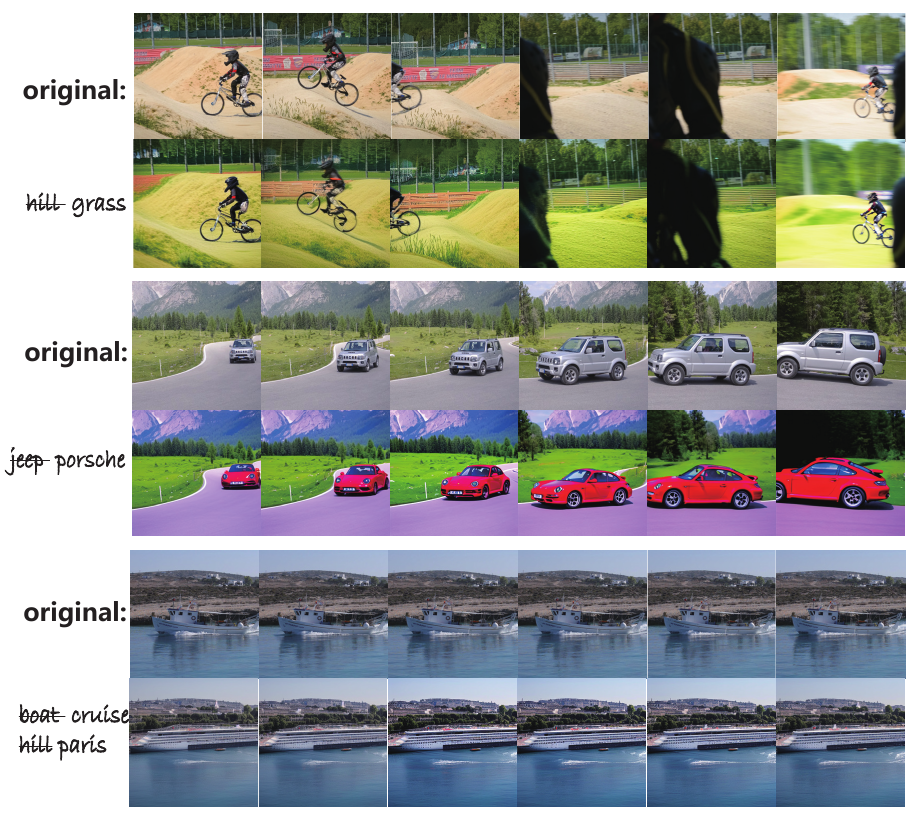}
\caption{\textit{RealCraft} enables zero-shot, shape-wise, consistent editing for real videos. Our method performs edits using Stable Diffusion, with text as the only input. No extra training or fine-tuning of models, structural guidance or parameter tuning is required.}
\label{fig:first_figure}
\vspace{-2em}
\end{figure}

Existing methods for text-driven real video editing often deploy image diffusion models, e.g., Stable Diffusion, for frame-wise edits while utilizing deterministic DDIM \cite{song2020denoising} for image-to-noise inversion. Three major methods are used to maintain the original layouts of the frames. One is to impose structural information such as Canny edges, human poses and bounding boxes utilizing ControlNet \cite{zhang2023adding}, thereby guiding the frame sampling process \cite{zhao2023controlvideo,jeong2023ground}. Another stream of work provides structural editing information through Atlas layer \cite{bar2022text2live,couairon2023videdit,chai2023stablevideo,lee2023shape} and various reference-frame-based methods \cite{yan2023magicprop,geyer2023tokenflow} are also deployed to maintain temporal consistency. In the aforementioned methods, additional inputs describing the structural information or supplementary models for generating intermediate data are usually needed and typically lack precise control over localized information, leading to challenges in compositional consistency \cite{he2023localized}. Attention-control-based methods inject new editing features during the denoising process by deploying the semantic relations in the cross-attention \cite{hertz2022prompt} or self-attention \cite{ceylan2023pix2video, wu2023tune} maps generated by pre-trained diffusion models. Using these approaches, zero-shot text-driven video editing is achieved \cite{qi2023fatezero}. However, these methods typically lack significant edits, e.g. shape and texture, over original frames or require precise tuning of parameters.

In this paper, we propose \textit{RealCraft}, a zero-shot method for text-driven shape-wise real video editing requiring no extra parameters. Our proposed method controls the attention maps generated by Stable Diffusion \cite{rombach2022high} and achieves a balance between significant localized edits and temporal consistency in edited videos. The method is built upon two key observations: i) An entire injection of cross-attention may cause instability across frames in real video editing. ii) Although spatial-temporal attention contributes to overall consistency, it overshadows new features injected by cross-attention.
 Based on the two observations, we design a two-step attention-control approach. Firstly, we propose to directly swap the cross-attention maps of the editing objects and inject the features during all DDIM denoising timesteps, thereby stabilizing the structure while localizing the edits. At the same time, we propose to relax the spatial-temporal attention's influence on new feature injection by masking out the feature-heavy areas, enabling significant shape edits while ensuring temporal consistency.

This two-step approach opens up the possibility for significant and consistent shape edits via swapping cross-attention for new feature injection and relaxing spatial-temporal attention for enhanced temporal consistency. This addresses a longstanding challenge in zero-shot real video editing, where normally a choice has to be made between these two aspects. The efficacy of the proposed method also alleviates attention-based methods' burden of careful parameter tuning, e.g., blending threshold, blending steps. The main contributions of our work are as follows.
\begin{itemize}
    \item We introduce \textit{RealCraft}, an effective method for zero-shot real video editing, featuring in significant shape editing ability with enhanced temporal consistency.
    \item We provide two key observations behind attention mechanism and propose a two-step approach of swapping cross-attention and relaxing spatial-temporal attention.
    \item We propose a localized cross-attention swapping technique. Specifically, it maintains the original prompt’s cross-attention after swapping, thereby ensuring stable feature injection.
    \item Qualitative and quantitative experiments have been conducted across various baselines, showcasing the efficacy of the proposed method on consistent video editing, background transformation, shape-wise editing, and pose preservation.
\end{itemize}

\section{Related Work}
\subsection{Diffusion Model for Text-driven Video Editing}
Diffusion Models (DMs) \cite{ho2020denoising} feature a step-wise denoising process, which facilitates enhanced control over sampling. Therefore, in recent years, DMs are widely studied for text-to-image generation  \cite{betker2023improving,saharia2022photorealistic,rombach2022high} and image editing \cite{sheynin2023emu,kawar2023imagic,parmar2023zero}.
 Text-driven video editing often takes advantages of the powerful pre-trained text-to-image models and adds effective control during the sampling process in preservation of structural layout and temporal consistency. ControlNet \cite{zhang2023adding} for conditional control in pre-trained diffusion models is a widely adopted approach \cite{ouyang2023codef,liao2023lovecon,yang2023rerender}. Diverse conditional controls are deployed, for example, HED boundaries \cite{zhao2023controlvideo}, various edge maps \cite{feng2023ccedit}, depth maps \cite{yan2023magicprop}, poses \cite{zhao2023make} and bounding boxes \cite{jeong2023ground}, catering to facial editing, structure maintenance, gesture preservation and target localization. However, this method requires choosing specified types of control for different video prompts and expertise in integration when multiple controls are needed. 

Neural Atlases (NLA)  model-based methods \cite{kasten2021layered,couairon2023videdit,lee2023shape,huang2023inve} are popular for their capacity to decompose inputs into layered representations, such as foreground and background atlases, ensuring scene consistency. Notable examples include Text2Live \cite{bar2022text2live}, which introduces additional edit layers to atlases and trains a specific generator for these layers. StableVideo \cite{chai2023stablevideo} enhances temporal consistency with an atlas aggregation network. However, the diversity of these methods is constrained by the object layer setup, or they require different pre-trained models for various videos.

Other methods include FuseYourLatent \cite{lu2023fuse}, which fuses latents directly during editing, and TokenFlow \cite{geyer2023tokenflow}, which propagates diffusion features using inter-frame correspondences. Dreamix \cite{molad2023dreamix}, on the other hand, corrupts the original video via down-sampling and focuses on fine-tuning for motion consistency, often at the expense of structural consistency.

\subsection{Attention Control for Image and Video Editing}
In addition to the aforementioned methods, attention-control-based approaches are becoming increasingly popular because they typically do not require additional input beyond text prompt and input video. Attention control for text-driven image and video editing generally involves two categories: self-attention control and cross-attention control. Since cross-attention can bridge between semantic information and image spatial layout, it has been widely adopted in image editing works \cite{wang2023dynamic,chen2023training,choi2023custom,park2023energy}. For instance, Prompt-to-Prompt \cite{hertz2022prompt} realizes localized image editing by replacing, refining, or reweighing the prompts' cross-attention weights. pix2pix-zero \cite{parmar2023zero} achieves structurally coherent image editing by using cross-attention as guidance. 
 
% When extending the editing target from images to videos, both spatial-temporal and cross-attention can be used to inject the features from the target prompt while maintaining temporal consistency. 
% FateZero\cite{qi2023fatezero} addressed the flickering issue in video editing by using the cross-attention of moving objects to mask the spatial-temporal attention, thereby minimizing background changes while injecting new features. P2P-video\cite{liu2023video} extend the cross-attention control in Prompt-to-prompt from image editing to video editing by adding sparse causal attention in the temporal domain. 
% LOVECon\cite{liao2023lovecon} processes the source video by splitting it into consecutive windows and sequentially editing the frames, while employing cross-attention between these windows to ensure temporal consistency. Pix2Video\cite{ceylan2023pix2video} conducts text-guided edits on a single frame and then propagates these changes to subsequent frames through self-attention injection.
When extending the editing target to videos, both self-attention and cross-attention can be effectively utilized to incorporate features from the target prompt while ensuring temporal consistency. Addressing the common flickering issue, Tune-A-Video~\cite{wu2023tune} introduced sparse-casual spatial-temporal attention to guarantee consistent generation. FateZero \cite{qi2023fatezero} leverages the cross-attention of moving objects to mask spatial-temporal attention through a user-specified blending threshold, minimizing background changes during feature injection. Video-P2P \cite{liu2023video} adapts the cross-attention control from Prompt-to-prompt, initially used in image editing, to video editing by integrating sparse-causal attention in the temporal domain. LOVECon \cite{liao2023lovecon} processes the source video by dividing it into consecutive windows, sequentially editing the frames while using cross-attention across these windows to maintain temporal consistency. However, maintaining strong temporal consistency often limits the freedom to introduce significant edits and thus restricts edits mainly to style transfer. While editing object shape by masking self-attention can potentially lead to instability if the blending threshold is not appropriately chosen. In this paper, we address these challenges by swapping cross-attention and relaxing spatial-temporal attention, allowing consistent shape-wise editing.

\section{Methodology}
\subsection{Preliminary}
\paragraph{\textbf{Latent Diffusion Models.} }Denoising Diffusion Probabilistic Models (DDPMs) \cite{ho2020denoising} are generative models that capture a data distribution $q(x)$ with a U-Net \cite{ronneberger2015u} $\epsilon_{\theta}$. Latent Diffusion Models (LDMs) \cite{rombach2022high} are variants of DDPMs in which the diffusing and denoising process are operated in the latent space of an autoencoder. An encoder $\mathcal{E}$ reduces an RGB image $x$ to a low-dimensional latent variable $z = \mathcal{E}(x)$, which is then approximated back to pixel space by decoder $\mathcal{D}$ as $\mathcal{D}(z) \approx x$. Text inputs can be incorporated into the model as guidance allowing semantic modifications of the input images. The objective of LDMs and text-driven LDMs is to minimize the losses, as shown in \textit{Eq.}~\ref{eq:ldm} and \textit{Eq.}~\ref{eq:text-driven ldm}.

\begin{equation}
    \mathcal{L}(\theta) = \mathbb{E}_{z_0\sim q(\mathcal{E}(x_{0})),\epsilon\sim\mathcal{N}(0,\; I),\; t} \left\|\epsilon - \varepsilon_\theta (z_t,\; t)\right\|^{2}_{2},
    \label{eq:ldm}
\end{equation}

\begin{equation}
    \mathcal{L}_{p}(\theta) = \mathbb{E}_{z_0\sim q(\mathcal{E}(x_{0})),\epsilon\sim\mathcal{N}(0,\; I),\; t} \left\|\epsilon - \varepsilon_\theta (z_t,\; t,\; p)\right\|^{2}_{2},
    \label{eq:text-driven ldm}
\end{equation}
where $p = \psi(P)$ is the embedding of the conditional text prompt $P$ and $z_t$ is a noisy sample of $z_0$ at timestep $t$, $\epsilon$ is a random Gaussian noise.

\paragraph{\textbf{DDIM Inversion and Sampling. }} Denoising diffusion implicit model (DDIMs) \cite{song2020denoising} are special cases of DDPMs where the sampling process is deterministic, and are thus employed for the inversion process for image and video editing tasks. The deterministic property not only accelerates the sampling process, but enables semantic interpolation in the latent variables whose consistency are ensured by deterministic sampling.

Random noise $z_T$ is converted to a clean latent $z_0$ in a sequence of timestep $t : T \rightarrow 1$:
\begin{equation}
    z_{t-1} = \sqrt{\alpha_{t-1}} \frac{z_{t} - \sqrt{1 - \alpha_{t}}\epsilon_{\theta}}{\sqrt{\alpha_{t}}} + \sqrt{1 - \alpha_{t-1}}\epsilon_{\theta},
    \label{eq:ddim_sampling}
\end{equation}

Grounded in the ODE limit analysis of the diffusion process, the DDIM inversion method is designed to transform a clean latent $\hat{z}_0$ into its corresponding noised version $\hat{z}_T$ through a reverse sequence of steps $t : 1 \rightarrow T$:

\begin{equation}
    \hat{z}_t = \sqrt{\alpha_{t}}\frac{\hat{z}_{t-1} - \sqrt{1 - \alpha_{t-1}}\epsilon_{\theta}}{\sqrt{\alpha_{t-1}}} + \sqrt{1 - \alpha_{t}}\epsilon_{\theta}.
    \label{eq:ddim_inversion}
\end{equation}
where $\alpha_t$ is the parameter for noise scheduling. 

\textit{Eq.}~\ref{eq:ddim_inversion} is employed to reverse the clean latent of the original video into its noised counterpart, followed by the application of \textit{Eq.}~\ref{eq:ddim_sampling} to sample the edited video from this noise space.
% sjin: ruiyu please help me check this part, what I wrote is slightly different from fatezero, i think they got logic error.

\begin{figure}
\centering
\includegraphics[height=12cm]{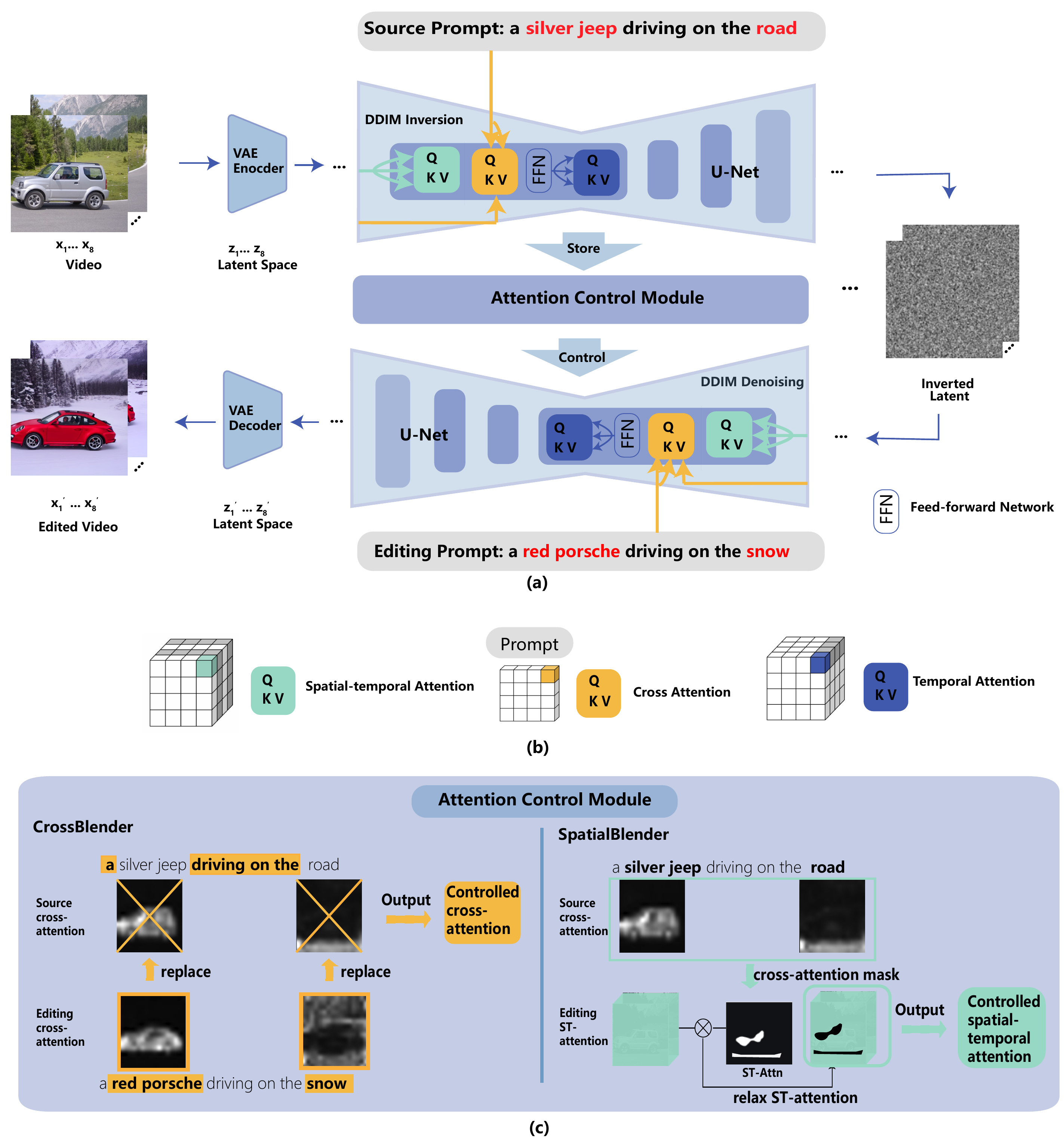}
\caption{(a) Our proposed \textit{RealCraft} pipeline takes source frames $\{{x}_{i}\}_{i=1}^{n}$ (n = 8 in this illustration), source prompt, and editing prompt as inputs. Initially, $\{{x}_{i}\}_{i=1}^{n}$ are encoded into latent space by a VAE \cite{kingma2013auto} encoder, followed by DDIM inversion to obtain the inverted latents, while storing spatial-temporal and cross-attention maps. In the denoising stage, the stored attention maps are fed into the Attention Control Module, orchestrating the spatial-temporal (\textsc{SpatialBlender}) and cross-attention (\textsc{CrossBlender}) for video editing. (b) Illustrations of spatial-temporal attention, cross attention, and temporal attention, with different colors representing the QKV components. Cross-attention occurs between the encoded prompt and frame. (b) The proposed Attention Control Module comprises \textsc{CrossBlender} and \textsc{SpatialBlender}.}
\label{fig:structure}
\vspace{-1.8em}
\end{figure}

\subsection{Attentions within \textit{RealCraft}}

% \subsubsection{Definitions. }In Stable Diffusion \cite{rombach2022high}, a U-Net architecture comprising 2D blocks with spatial self-attention and cross-attention layers is employed for DDIM inversion (\textit{Eq.}~\ref{eq:ddim_inversion}) and denoising (\textit{Eq.}~\ref{eq:ddim_sampling}), with spatial-attention catering to consistency within the frame and cross-attention (Cross-Attn) aligning frame content and editing prompt. Tune-A-Video \cite{wu2023tune} inflates those 2D blocks into 3D, converting self-attention into spatial-temporal attention (ST-Attn) to enable video generation. The definition of the content-decisive $\textsc{Cross-Attn}(Q_c,\;K_c,\;V_c)$ and $\textsc{ST-Attn}(Q_{st},\;K_{st},\;V_{st})$ is as follows:
In Stable Diffusion \cite{rombach2022high}, a U-Net architecture comprising 2D blocks with spatial self-attention and cross-attention layers is employed during DDIM inversion and denoising process for text-to-image generation. Specifically, cross-attention aligns the image spatial layout with a text prompt, while self-attention provides consistency within the image. In \textit{RealCraft}, we utilize cross-attention between embedded frames $[z_{t}]_{t=1}^{T}$ and encoded editing prompts $p$ to inject editing features and follow the design of Tune-A-Video \cite{wu2023tune} to inflate 2D self-attention into 3D spatial-temporal attention, enabling video editing. $\textsc{Cross-Attn}(Q_c,\;K_c,\;V_c)$ and $\textsc{ST-Attn}(Q_{st},\;K_{st},\;V_{st})$ are defined as follows.

\begin{equation}
    Q_c=W^{Q_c} \cdot z_t^{k}, K_c=W^{K_c} \cdot p, V_c=W^{V_c} \cdot p
    \label{eq:Cross-Attn}
\end{equation}
% \vspace{-1em}
\begin{equation}
    Q_{st}=W^{Q_{st}} \cdot z_{t}^{k}, K_{st}=W^{K_{st}} \cdot [z_{t}^{1},\;z_{t}^{k-1}], V_{st}=W^{V_{st}} \cdot  [z_{t}^{1}, \;z_{t}^{k-1}]
    \label{eq:ST-Attn}
\end{equation}

where the $k \in [1,\; K]$, $K$ is the number of total frames to edit, $z_t^k$ represent the frame $k$ in latent space at timestep $t$; $W^{Q_c}$, $W^{K_c}$, $W^{V_c}$, $W^{Q_{st}}$, $W^{K_{st}}$, $W^{V_{st}}$ are projection matrices from the pre-trained model \cite{rombach2022high}. An illustration of types of Attention used are shown in \textit{Fig.}~\ref{fig:structure} (b). 

As illustrated in \textit{Fig.}~\ref{fig:structure}, during the inversion process, all blocks of cross-attention maps for each word within the source prompt and spatial-temporal attention maps are stored. This storing concept was introduced by FateZero \cite{qi2023fatezero} to enable zero-shot editing. We use the stored source attention maps to facilitate cross-attention swapping and spatial-temporal attention relaxing for consistent new feature injection during the denoising stage with the proposed Attention Control Module.
% Edits are executed during the denoising stage, orchestrated by the Attention Control Module, with \textsc{CrossBlender} catering to consistent editing and \textsc{SpatialBlender} avoiding modifications made by \textsc{CrossBlender} being overshadowed.

% Drawing on FateZero's \cite{qi2023fatezero} concept of storing attention for editing, we introduce the following notations for clarity: 
% \[P_{src} = \{[w^{src}]_{0}, [w^{src}]_{1},\ldots,[W^{src}]_{n-1}\},\: P_{edit} = \{ [w^{edit}]_{0}, [w^{edit}]_{1}, \dots,
% [W^{edit}]_{n-1}\}\] 
% \[W_{src} = \{[W^{src}]_{0}, [W^{src}]_{1}, \dots, [W^{src}]_{N-1}\}\: W_{edit} = \{ [W^{edit}]_{0}, [W^{edit}]_{1}, \dots, [W^{edit}]_{N-1}\}\]
Some notations are introduced for clarity:
$P^{src} = \{p^{src}_{i} \mid i \in [1,\; M]\}$, $P^{edit} = \{p^{edit}_{i} \mid i \in [1,\; M]\}$ denote the source and editing prompts, $M$ is the total number of words in the prompt. $W^{src} = \{w^{src}_{j} \mid j \in [1,\; N]\}$, $W^{edit} = \{w^{edit}_{j} \mid j \in [1,\; N]\}$ represent the words to be modified in $P^{src}$ and the replacement words in $P^{edit}$, where $N$ is the number of words to edit. At timestep $t$ during denoising, the cross-attention maps $c^{edit}_{i,\; t} \in C^{edit}_{t}$ for each element in $P^{edit}$ and the spatial-temporal attention map $s^{edit}_{t}$ are calculated. Correspondingly, the stored cross and spatio-temporal attention maps of source prompts during the inversion are $[C^{src}_{t}]^{T}_{t=1} = \{c^{src}_{i,\; t} \mid i \in [1,\; M]\}$ and $[s_{t}^{src}]_{t=1}^{T}$.

\subsection{Attention Control Module in \textit{RealCraft}}
The Attention Control Module comprises two key components: \textsc{CrossBlender} and \textsc{SpatialBlender}, tasked with swapping cross-attention and relaxing spatial-temporal attention, respectively. 

\subsubsection{CrossBlender}
% Prompt-to-prompt \cite{hertz2022prompt} introduced an effective image editing method by fusing cross-attention layers for each textual token. Our \textsc{CrossBlender}, while inspired by the swapping method in Prompt-to-Prompt, adopts a distinct approach. It diverges from Prompt-to-Prompt as follows: 1. We swap only the cross-attention of specified words ($W^{src}$ and $W^{edit}$), rather than the entire cross-attention map of the sentence 2. Unlike the step-function-like swapping in Prompt-to-Prompt, which is determined by the timestep, our method replaces the cross-attention of $W^{src}$ directly with that of $W^{edit}$, independent of the timestep, thus realized parameter-free editing. 3. After swapping, we use the cross-attention of $P^{src}$ as the controlled output, rather than retaining the cross-attention of $P^{edit}$. These differences result in more localized changes by swapping the attention of specific words only; a complete injection ensures greater accuracy of the inserted features without the need for tuning timestep; and a focus on the temporal consistency of the generated frames. The nature of the step-function-like fusing process makes prompt-to-prompt primarily designed for synthetic images. When Video-P2P \cite{liu2023video} transferred this fusing technique to videos, as shown in \textit{Fig.}~\ref{fig:applications}, the edits are splatted over the frame, leading to inconsistency and problematic poses.
Prompt-to-prompt \cite{hertz2022prompt} introduces cross-attention maps for feature injection in text-driven image editing. Specifically, the entire source prompt's cross-attention maps are injected during the DDIM denoising, and an injection interval needs to be decided. We observe that this step-function-like injection does not accommodate video editing's consistency requirement, as structural instability can be introduced when edits are not localized, examples can be found in \textit{Fig.}~\ref{fig:applications}. Our \textsc{CrossBlender} adopts a distinct approach in swapping the cross-attention maps: i) Rather than directly injecting the entire source prompt maps, we swap the cross-attention map of the editing target (from $W^{src}$ to $W^{edit}$), which implies the cross-attention injected during the denosing is a combination of source maps $[C_{t}^{src}]_{t=1}^{T}$ stored during the DDIM inversion process and new maps calculated by \textit{Eq.} \ref{eq:Cross-Attn} for $W^{edit}$. This localizes changes by swapping specific words, resulting in new feature injection along with minimal influence on the structural layout of each frame. ii) Instead of injecting new features from certain timestep, our \textsc{CrossBlender} outputs the swapped cross-attention maps during the whole denoising process. This complete injection provides a stable source of new features and alleviates the need to choose a specific injecting interval. 
% The nature of the step-function-like fusing process makes prompt-to-prompt primarily designed for synthetic images. When Video-P2P \cite{liu2023video} transferred this fusing technique to videos, as shown in \textit{Fig.}~\ref{fig:applications}, the edits are splatted over the frame, leading to inconsistency and problematic poses.
The mathematical representation of \textsc{CrossBlender} at timestep $t$ is as follows.
\begin{equation}
    \textsc{CrossBlender}(C^{src}_{t},\; C^{edit}_{t}) = \{c_{i,\; t}\}_{i=1}^{M} = \begin{cases}
c^{src}_{i,\; t} & \text{ for } p^{edit}_{i} \text{ not in } W^{edit}\\ 
c^{edit}_{i,\; t} & \text{ for } p^{edit}_{i} \text{ in } W^{edit}
\end{cases}
    \label{eq:crossblender}
\end{equation}

% \begin{figure}[h]
% \vspace{-1.8em}
% \centering
% \includegraphics[height=1.9cm]{figures/heatmap_v1.pdf}
% \caption{A demonstration of the impact of blending threshold $\tau$ on blending mask. }
% \label{fig:heatmap}
% \end{figure}

\subsubsection{SpatialBlender}
% \begin{figure}
% \vspace{-1.8em}
% \centering
% \includegraphics[height=1.9cm]{figures/heatmap_v1.pdf}
% \caption{A demonstration of the impact of blending threshold $\tau$ on blending mask. }
% \label{fig:heatmap}
% \end{figure}

Another key observation from us is that spatial-temporal attention, although contributes to overall temporal consistency, can largely weaken the injection of new features through cross-attention. As a result, the edited frames often closely resemble the source frames in shape, color and texture, making the modifications on objects and background less noticeable, examples can be found in \textit{Fig.}~\ref{fig:background_transformation} and \textit{Fig.}~\ref{fig:applications}. This is partially because the DDIM inversion process involves a trade-off between distortion and editability \cite{tov2021designing}, where enhancing reconstruction by reducing prompt influence limits the capacity for significant edits \cite{hertz2022prompt}. We find that by relaxing the spatial-temporal attention in the feature-heavy area, significant shape edits with enhanced temporal consistency can be made. 
%along with the specific blending steps for self-attention ($t_s$) and cross-attention ($t_c$). As the room newly opened from relaxation needs to be filled with a more stable source of new features rather than adding the spatial-temporal attention of original background and editing object with a binary mask. 

\begin{figure}[h]
\vspace{-1em}
\centering
\includegraphics[height=1.9cm]{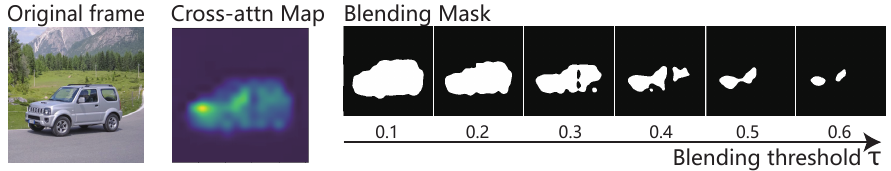}
\caption{A demonstration of the impact of blending threshold $\tau$ on blending mask. }
\label{fig:heatmap}
\end{figure}

Relaxation here refers to masking out feature-heavy areas in the original spatial-temporal attention $s_{t}^{src}$, and replacing the rest with the editing spatial-temporal attention $s_{t}^{edit}$. The feature-heavy area can be extracted from a blending mask $M_{t}$, generated with \textit{Eq.} \ref{eq:blending_mask} using stored cross-attention maps $C_{t}^{src}$ with fixed $\tau$ equals to 0.5. Conversely, FateZero \cite{qi2023fatezero} proposes to mask out the edited area to keep the original structure unchanged, with the same \textit{Eq.} \ref{eq:blending_mask} but user-specified $\tau$. However, as shown in \textit{Fig.}~\ref{fig:heatmap}, the shape of $M_{t}$ varies a lot with different $\tau$. This could be why FateZero's performance is highly dependent on the careful selection of the threshold value to prevent flickering issues. This is due to the fact that, even though spatial-temporal attention limits significant edits, it still contributes to the overall consistency. Thus when relaxing the spatial-temporal attention, a higher consistency requirement for the features injected by cross-attention is raised. And \textsc{CrossBlender} is able to provide a more stable source of features rather than solely relying on the cross-attention between inverted latent and the editing prompt. The mathematical representation of \textsc{SpatialBlender} at timestep $t$ is as follows.
\begin{equation}
    M_{t} = \textsc{Step}(C^{src}_{t},\; 0.5)
    \label{eq:blending_mask}
% \vspace{-1.2em}
\end{equation}
\begin{equation}
\textsc{SpatialBlender}(C^{src}_{t},\; s_{t}^{src},\; s_{t}^{edit}) = M_{t}\odot s_{t}^{edit} + (1 - M_{t})\odot s_{t}^{src}
    \label{eq:spatialblender}
\end{equation}
where $\textsc{Step}$ is a Heaviside step function with a constant threshold equals to 0.5. Note that spatial blending only works for editing words $W^{edit}$.

% \begin{figure}
% \vspace{-1.8em}
% \centering
% \includegraphics[height=1.9cm]{figures/heatmap_v1.pdf}
% \caption{A demonstration of the impact of blending threshold $\tau$ on blending mask. }
% \label{fig:heatmap}
% \end{figure}

\subsubsection{Algorithm}
 The general algorithm of \textit{RealCraft} video editing with \textsc{CrossBlender} and \textsc{SpatialBlender} is shown in \textit{Algorithm}~\ref{al:realcraft}.
 
\begin{algorithm}[h]
\caption{\textit{RealCraft} video editing}
\label{al:realcraft}
\begin{algorithmic}[1]
\State \textbf{Input:} original video $X$, source prompt $P^{src}$, editing prompt $P^{edit}$
\State \textbf{Output:}  Edited video $X^{edit}$
\State $z_0 \leftarrow \mathcal{E}(X)$;
\State $z_{T}$,\; $[C^{src}]^{T}_{t=1}$,\; $[s^{src}]_{t=1}^{T}$ = $\textsc{DDIM\_INV}(z_0,\; P^{src})$;
\State $z_{T}^{*} \leftarrow z_{T}$;
\For{$t = T;\; T - 1,\; \dots,\; 1$}
    \State $z_{t-1},\; C^{edit}_{t-1},\; s_{t-1}^{edit} \leftarrow \textsc{DDIM}(z_t,\; P^{edit},\; t)$;
    \State $C^{edit*}_{t-1} \leftarrow \textsc{CrossBlender}(C^{src}_{t-1},\; C^{edit}_{t-1})$;
    \State $s_{t-1}^{edit*} \leftarrow \textsc{SpatialBlender}(C^{src}_{t-1},\; s_{t-1}^{src},\; s_{t-1}^{edit})$;
    \State $z_{t-1}^* \leftarrow \textsc{DDIM}(z_{t}^{*},\; P^{edit},\; t)\{C^{edit}_{t-1} \leftarrow C^{edit*}_{t-1},\; s_{t-1}^{edit} \leftarrow s_{t-1}^{edit*} \}$;
\EndFor
\State $X^{edit} \leftarrow \mathcal{D}(z_{0}^*)$
\State \textbf{return} $X^{edit}$
\end{algorithmic}
\end{algorithm}

\section{Experiments}
\subsection{Implementation Details}
For most of the editing tasks, we directly employ \textit{stable-diffusion-v1-4} \cite{Rombach_2022_CVPR} as the pre-trained model with a total DDIM timestep of $T=30$. In the quantitative evaluation, to ensure fair evaluation and accommodate baselines requiring atlas layers, whose extraction requires training of models for each video, we utilize the dataset in NLA \cite{kasten2021layered}, sourced from the DAVIS dataset \cite{pont20172017}, comprising real videos ranging from 43 to 70 frames. These frames are further segmented into non-overlapping sets of eight consecutive frames, each paired with 2 or 3 editing prompts, to conduct the evaluation. Across all baselines, where applicable, a uniform blending threshold of 0.5,  fixed blending steps $t \in [0.5 T,\; T]$, pre-trained atlas layer sourcing from NLA \cite{kasten2021layered} are adopted to ensure fair evaluation.

% \begin{figure}
% \centering
% \includegraphics[height=3.2cm]{figures/background_transformation_v5.pdf}
% \caption{Qualitative comparison with other baselines in background transformation. }
% \label{fig:background_transformation}
% \vspace{-2em}
% \end{figure}

\subsection{Metrics} 
% we added more, re-write this part
Our user study evaluated the editing quality, editing consistency, and edited video fidelity across 64 videos with 29 subjects. The subjects were presented with the original video and 7 anonymous edited videos at the same time, then asked to grade each video on a scale from 1 (Very bad) to 5 (Very good). Additionally, we use the same quantitative metrics adopted by the FateZero\cite{qi2023fatezero}, the metrics are as follows:
\begin{itemize}
    \item \textbf{Tem-Con \cite{esser2023structure}.} This metric assesses temporal consistency in video frames by calculating the cosine similarity across all pairs of consecutive frames.
    \item \textbf{Frame-Acc \cite{parmar2023zero}. }This refers to frame-wise editing accuracy, quantified as the proportion of frames where the edited image exhibits higher CLIP similarity \cite{radford2021learning} to the editing prompt compared to the source prompt.
    % \item \textit{Warp error} measures temporal consistency by comparing the mean squared error between warped edited frames and target frames, reflecting pixel-level accuracy. 
    \item \textbf{Edit (User Study).} This evaluates the quality of edits, focusing on how well they align with the prompt, and the suitability of shape, background, and object features.
    \item  \textbf{Temp (User Study). } Edited video's temporal consistency and its adherence to the original video's sequence.
    \item \textbf{Fidelity (User Study).} Average score representing participants' perceived video fidelity ranging from artificial (1) to real (5).
\end{itemize}

\subsection{Baseline Comparisons}
Quantitative evaluations of our proposed method in comparison with state-of-the-art zero-shot video editing baselines are performed. We select 6 baselines, each as a representative of one category within the spectrum of video editing methodologies, covering three mainstream approaches outlined in Section 2: attention-control-based, ControlNet, and NLA-based methods. i) FateZero~\cite{qi2023fatezero} conducts editing through masking the spatial-temporal attention with binary cross-attention masks. ii) Tune-A-Video~\cite{wu2023tune} generates similar content by overfitting an inflated diffusion model on a single video. iii) TokenFlow \cite{geyer2023tokenflow} leverages the original inter-frame feature correspondences when editing features to achieve better temporal consistency. iv) LOVECon~\cite{liao2023lovecon} adopts ControlNet~\cite{zhang2023adding} to guide the denoising process, employing cross-attention across frames' windows to facilitate long video editing. v) StableVideo~\cite{chai2023stablevideo} performs consistent shape-aware video editing through updating the models Neural layered atlas (NLA) via editing key video frames and vi) Video-P2P~\cite{liu2023video} expands Prompt-to-Prompt~\cite{hertz2022prompt}'s cross-attention control from image to video editing by incorporating spatial-temporal attention.

\begin{table}[ht]
\vspace{-1em}
\centering
\caption{Quantitative evaluation against baselines on dataset provided in NLA \cite{kasten2021layered}.}
\label{tab:long_video}
\begin{tabular}{cccccccc}
\toprule
\multicolumn{1}{c}{Method} & \multicolumn{2}{c}{CLIP Metrics$\uparrow$} & \multicolumn{1}{c}{} & \multicolumn{3}{c}{User Study$\uparrow$}  \\
\cmidrule{2-3} \cmidrule{5-7} 
 & Tem-Con & Frame-Acc && Edit & Temp & Fidelity\\
\hline
FateZero \cite{qi2023fatezero} & {0.9564} & {\underline{0.9043}} & {} &{2.26} & {2.11}& {2.41}\\
Tune-A-Video \cite{wu2023tune} & {0.9718} & {0.8544} & {} & {3.26} & {3.37} & {3.33} \\
TokenFlow \cite{geyer2023tokenflow} & {0.9303} & {0.7574}  & {} & {\underline{3.41}} & {\textbf{4.26}} & {\underline{3.81}}\\
LOVECon \cite{liao2023lovecon} & {0.8859} & {0.6545}  & {} & {1.89} & {1.01} & {1.44} \\
StableVideo \cite{chai2023stablevideo} & {0.9238} & {0.8469}  & {} & {1.67} & {3.00} & {1.78} \\
Video-P2P \cite{liu2023video} & {0.8798} & {0.8689}  & {} & {2.48} & {2.00} & {1.56}\\
\hline
\textbf{Ours} - \textit{C. only} & {\underline{0.9727}} & {0.8700}  & {} & {--} & {--} & {--}  \\
\textbf{Ours} - \textit{C. + S.} & {\textbf{0.9774}} & {\textbf{0.9542}}  & {} & {\textbf{3.74}} & {\underline{3.74}} & {\textbf{3.93}} \\
\bottomrule
\end{tabular}
\begin{tablenotes}    
    \footnotesize 
        \item[1.] \textit{C. only} stands for only \textit{CrossBlender} is used.
        \item[2.] \textit{C. + S.} stands for both \textit{CrossBlender} and \textit{SpatialBlender} are used.
        \item[3.] The highest performances are highlighted in bold, the second-best results are marked with an underline.
\end{tablenotes}
\end{table}

% \begin{figure}
% \centering
% \includegraphics[height=3.2cm]{figures/background_transformation_v5.pdf}
% \caption{Qualitative comparison with other baselines in background transformation. }
% \label{fig:background_transformation}
% \vspace{-2em}
% \end{figure}

\subsection{Experimental Results}
\textit{Tab.}~\ref{tab:long_video} displays the performance of our method and various baselines. Our approach outperforms others in both metrics. ``Tem-Con'' measures the temporal consistency across edited frames while ``Frame-Acc'' evaluates the edited video's alignment with the editing prompt, reflecting whether significant changes have been made in the edits.  In our user study, our method was rated highest for edit quality and fidelity, and second for temporal consistency, with TokenFlow being the top in that category.

\begin{figure}[h]
\centering
\includegraphics[height=3.2cm]{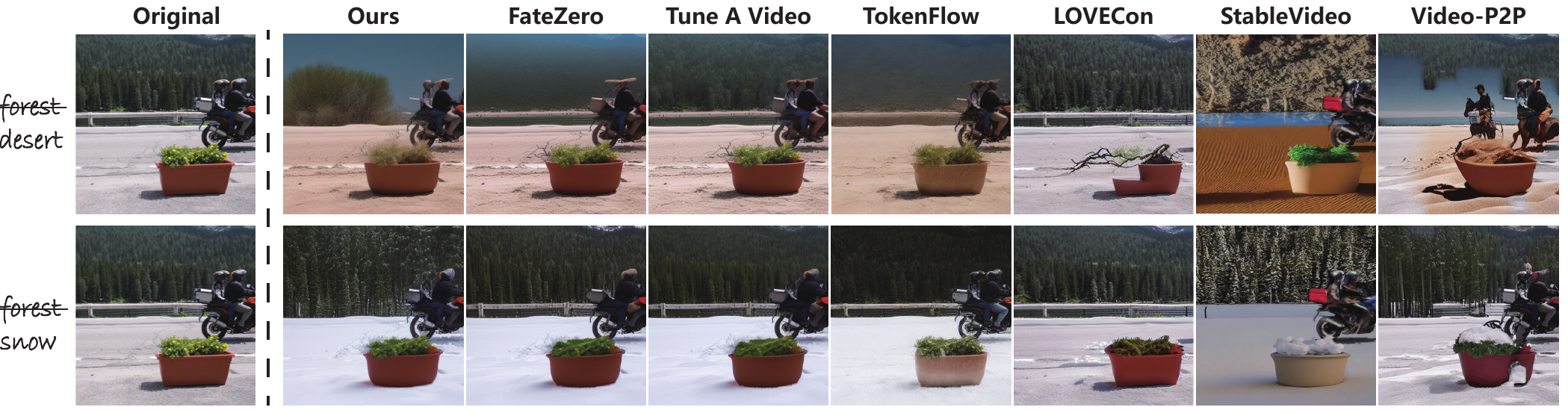}
\caption{Qualitative comparison with other baselines in background transformation. }
\label{fig:background_transformation}
% \vspace{-2em}
\end{figure}

\begin{figure}
\vspace{-1em}
\centering
\includegraphics[width=1\textwidth]{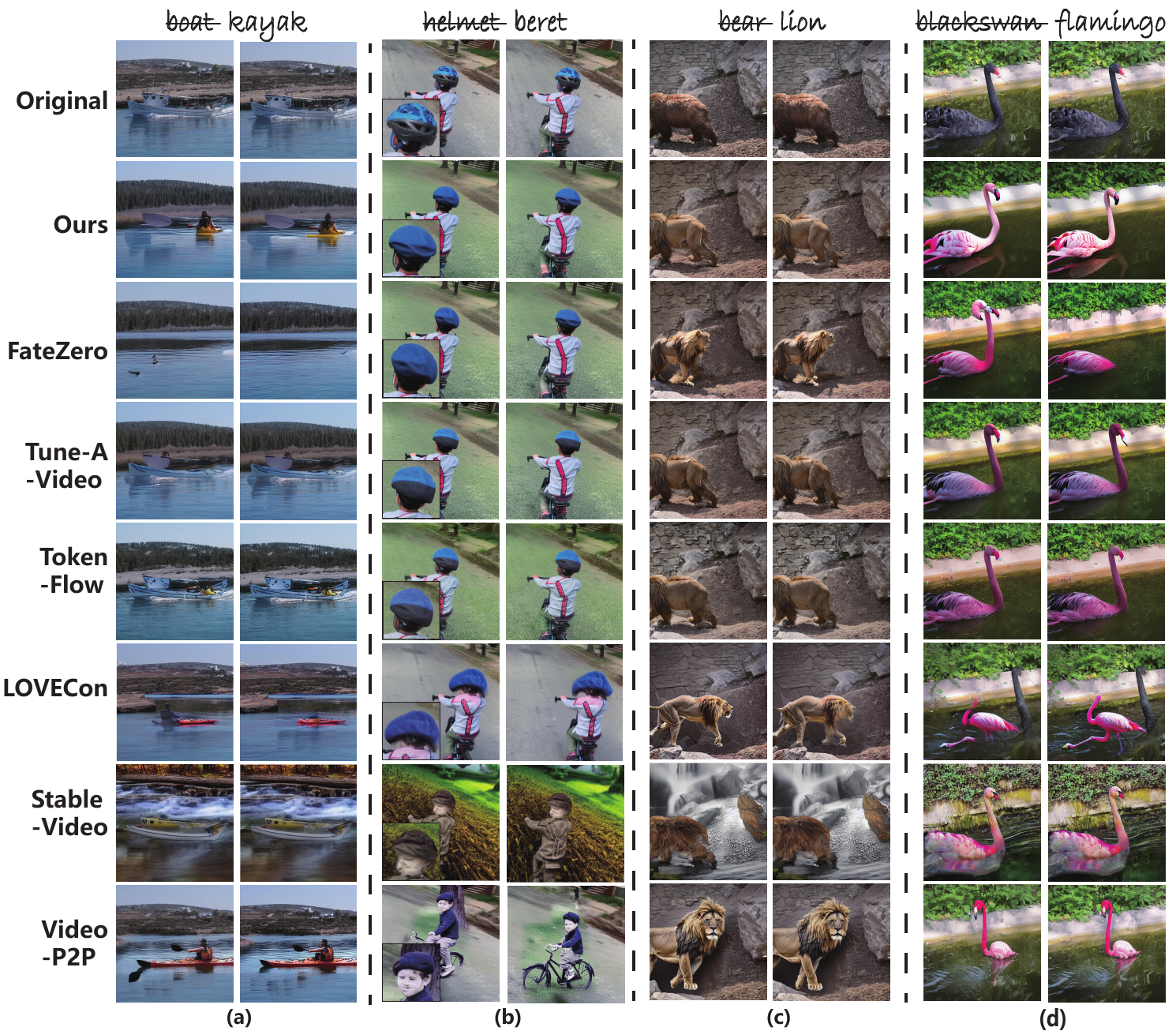}
\caption{Qualitative comparison with other baselines in shape editing: (a) $boat \rightarrow kayak$ and $hill \rightarrow forest$; (b) $helmet \rightarrow beret$ and $road \rightarrow grass$, and pose preservation: (c) $bear \rightarrow lion$; (d) $blackswan \rightarrow flamingo$}
\label{fig:applications}
\vspace{-1em}
\end{figure}

\textit{Fig.}~\ref{fig:background_transformation} - \ref{fig:applications} display the qualitative comparison. In \textit{Fig.}~\ref{fig:background_transformation}, significant background modifications are made to align with editing prompts, such as transforming forests into deserts or snowy landscapes, with \textit{RealCraft} adding specific features like desert terrain and snow on trees, a feature rarely presented by the others. 

\textit{Fig.}~\ref{fig:applications} showcases the selected methods' proficiency in shape editing and pose preservation. Full frames and more examples can be found in the supplementary material. In (a), \textit{RealCraft} and Video-P2P transform the boat into a kayak with significant shape changes. FateZero struggles in adding the new object, likely due to sensitivity in selecting the optimal blending threshold. Even though Tune-A-Video, TokenFlow, LOVECon and StableVideo convert the boat to a kayak, the shape of the kayak still follows the original structure of the boat, which results in reduced video fidelity. We observe similar results in (b), where some methods appear less effective in capturing a realistic shape and texture of the beret, with the color distribution remaining similar to the original helmet.

% This can also be seen in the helmet-to-beret example, some methods fall short in achieving a realistic shape and texture, with the color distribution closely mirroring the original frames. 

% \textit{Fig.}~\ref{fig:pose}(a) illustrates the ability of various methods to preserve pose, highlighting the challenge in real videos where objects may not be fully visible, different from synthesized videos. When preserving pose, it's important to localize edits while allowing for subtle adjustments. For example, in blackswan-to-flamingo, it's essential not just to match the flamingo's position to the blackswan's but also to make adjustments that enhance the resemblance to a flamingo, such as altering neck curvature, which is neglected in some methods. In the jeep-to-porsche example, we achieve precise shape editing while maintaining the original pose.
% \textit{Fig.}~\ref{fig:pose} (a) and illustrates the ability of various methods to preserve pose where objects are not fully visible, our method, Tune-A-Video, TokenFlow and StableVideo successfully maintain the original pose. In \textit{Fig.}~\ref{fig:pose}(b), when transforming the blackswan to flamingo, it's essential not just to match the position but also to allow subtle adjustments that enhance the resemblance to a flamingo, such as altering neck curvature, which is neglected in some methods. In the jeep-to-porsche example, we achieve precise shape editing while maintaining the original pose.
\textit{Fig.}~\ref{fig:applications} (c) compares the models' ability in pose preservation, the information of which is not fully observable from frame and text. FateZero, LOVECon and Video-P2P fail to maintain the original pose and orientation of the lion. Furthermore, our method ensures higher editing fidelity as shown in (d). When transforming the blackswan to flamingo, alternation of the neck curvature of the swan leads to enhanced fidelity, which is not conveyed in some methods. This serves as a supplementary example of a successful shape edit: the design of Attention Control Module appears to allow adequate new feature injection in this example.

% \begin{figure}
% \centering
% \includegraphics[height=3.2cm]{figures/background_transformation_v5.pdf}
% \caption{Qualitative comparison with other baselines in background transformation. }
% \label{fig:background_transformation}
% \vspace{-2em}
% \end{figure}

% \begin{figure}
% \vspace{-3em}
% \centering
% \includegraphics[width=0.9\textwidth]{figures/applications.pdf}
% \caption{Qualitative comparison with other baselines in shape editing: (a) $boat \rightarrow kayak$ and $hill \rightarrow forest$; (b) $helmet \rightarrow beret$ and $road \rightarrow grass$, in pose preservation: (c) $bear \rightarrow lion$; (d) $blackswan \rightarrow flamingo$}
% \label{fig:applications}
% \vspace{-2em}
% \end{figure}

It can be concluded that while the attention-controlled-based FateZero, LOVECon and Video-P2P succeed in introducing new features, flickering issues are often observed, likely due to the sensitivity of blending threshold and timesteps. Among them, Video-P2P also faces challenges in preserving the original structural layout across frames, suggesting more localized control instead of step-function-like swapping over cross-attention is needed for videos. Tune-A-Video and TokenFlow maintain strong temporal consistency, yet the edits often resemble the original frames in terms of color and shape. The NLA-based StableVideo demonstrates good performance in both shape editing and temporal consistency, however, the fidelity issues, potentially due to the layer-wise editing, are raised in the user study. Experimental results indicate that our method improves the editing performance by balancing significant edits and temporal consistency. 

% Prioritizing temporal consistency sometimes may cause spatial-temporal attention to overshadow cross-attention. This often leads to edits focusing mainly on the ground, with unchanged color tones and textures from the original frames, which can be seen in the forest-to-snow examples

% \begin{figure}
% \centering
% \includegraphics[height=9cm]{figures/shape_v3.pdf}
% \caption{Qualitative comparison with other baselines in shape editing: (a) $boat \rightarrow kayak$ and $hill \rightarrow forest$; (b) $blackswan \rightarrow duck$; (c) $helmet \rightarrow beret$ and $road \rightarrow grass$ }
% \label{fig:shape_editing}
% \vspace{-1em}
% \end{figure}
\subsection{Applications}

\subsubsection{Consistent Video Editing.}
\begin{figure}[h]
\centering
\includegraphics[height=3.5cm]{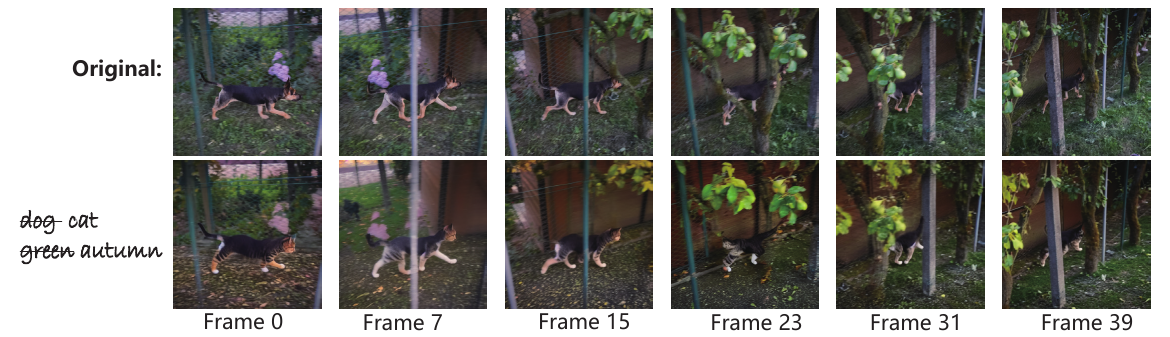}
\caption{An example of \textit{RealCraft} applied to a challenging video comprising a fast-moving object and occlusions, spanning 40 frames.}
\label{fig:consistency}
\vspace{-1em}
\end{figure}

\textit{RealCraft} supports temporal-consistent video editing up to 64 frames using a sliding window approach. It processes 8 frames at a time without any overlap or prior information on previous frames. An example is shown in \textit{Fig.}~\ref{fig:consistency}, where the original video features a small dog moving quickly under and with occlusions caused by leaves, poles, and fences. Our method transforms the dog into a cat while preserving the video's overall temporal consistency. Another example can be found in \textit{Fig.}~\ref{fig:first_figure} hill-to-grass, the edits remain stable with fast-moving objects and severe occlusions.
% \textit{RealCraft}'s consistency stems from its localized edits, which are controlled by the object-centric cross-attention swapping.
\vspace{-1em}

\subsubsection{Background Transformation.} When performing the background transformation with attention-control-based methods, we observe that the edited background closely resembles the original one in terms of texture and color distribution. Leveraging the relaxation of spatial-temporal attention and the swapping of cross-attention, \textit{RealCraft} neutralizes the influence of the original structure while introducing new features, preventing them from being overshadowed by existing elements in original frames. Examples can be found in \textit{Fig.}~\ref{fig:first_figure} and \textit{Fig.}~\ref{fig:background_transformation}. We observe that in our experiments, our approach ensures clean and consistent background transformation. 
\vspace{-1em}

\subsubsection{Precise Shape Editing.} To maintain temporal consistency in video editing, existing methods' edits on objects primarily focus on texture, instead of shape \cite{lee2023shape}. As a result of the aforementioned two-step approach, \textit{RealCraft} enables consistent shape editing. As shown in \textit{Fig.}~\ref{fig:applications} (a) and (b), the shape of the original objects is altered to closely match the target's features while maintaining the edits localized and stable across frames.
\vspace{-1em}

% \begin{figure}
% \centering
% \includegraphics[height=3.5cm]{figures/time_consistency_v2.pdf}
% \caption{An example of \textit{RealCraft} applied to a challenging video comprising a fast-moving object and occlusions, spanning 40 frames.}
% \label{fig:consistency}
% \vspace{-1em}
% \end{figure}

\subsubsection{Pose preservation.} As mentioned by previous research \cite{qi2023fatezero, wang2023zero}, preservation of an object's pose is crucial to ensure fidelity and temporal consistency. The difficulty increases when the objects described by prompts are not fully visible. For instance, in \textit{Fig.}~\ref{fig:applications} (c), despite the bear's head being left out from the image – a situation that typically complicates editing because distinct features are needed by the prompts to provide enough guidance – \textit{RealCraft} is able to retain the original pose. In \textit{Fig.}~\ref{fig:applications} (d), preservation of the original poses and reasonable shape modifications are carried out at the same time, this ability stems from the localized edits made by cross-attention swapping.

\subsection{Ablation Study}
\begin{figure}[h]
\vspace{-1em}
\centering
\includegraphics[height=4cm]{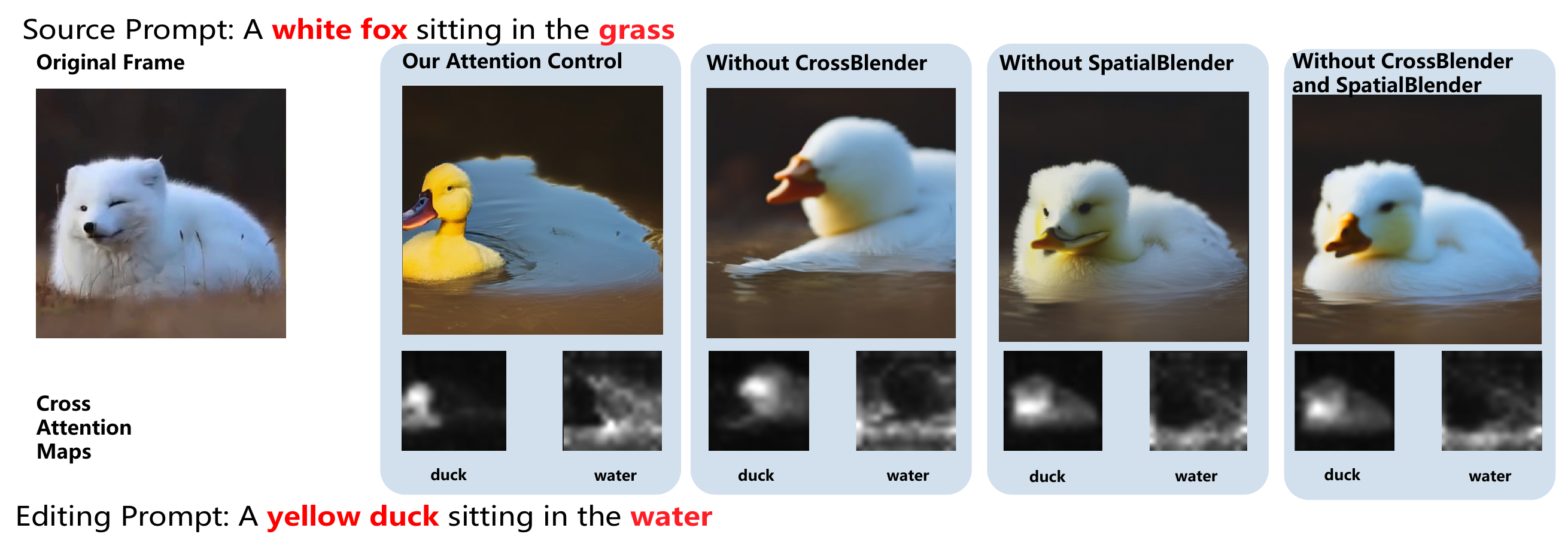}
\caption{An ablation study on the Attention Control Module's components for shape editing: The top row displays the original and the generated frames, while the second row illustrates the cross-attention maps for the editing words $W^{edit}$. }
\label{fig:duck}
\vspace{-1em}
\end{figure}

An ablation study is performed on the 2 types of attention-control components to illustrate of their individual contributions to model performance. As shown in \textit{Fig.}~\ref{fig:duck}, the object white fox is changed to yellow duck and the background transforms from grass to water. With both \textsc{CrossBlender} and \textsc{SpatialBlender}, the vacant areas in the background caused by significant changes are automatically integrated with water. Without \textsc{CrossBlender}, the edits fail to be localized and lead to low editing fidelity. While the absence of \textsc{SpatialBlender} weakens the introduction of the new features, thus resulting in minor change of shape and color. When both are missing, the edited results completely fail to align with the editing prompt's requirement of ``yellow''.

% \section{Discussion}
% \subsection{Failure Case}
% Our method depends on accurate cross-attention masks between the original frame and prompt provided by Stable Diffusion. In \textit{Fig.}~\ref{fig:failure_case}, a failure case is depicted, the transform of a gray dog to a robotic dog with the mat to floor is incomplete. This issue arises because the cross-attention mask does not entirely align with the ground truth, such as the attention mask for the mat focusing mainly on the edges and overlooking the central part, thereby limiting edits to the attention-focused areas. However, fine-tuning Stable Diffusion for specific downstream tasks can alleviate this problem.
% \begin{figure}
% \centering
% \includegraphics[height=3cm]{figures/failure_case.pdf}
% \caption{An illustration of a failure case: (a) shows the original frame, (b) depicts the edited frame where 'gray' dog is changed to 'robotic' dog with 'mat' to 'floor', and (c) displays the cross-attention masks between the prompt words('gray', 'mat') with the original frame.}
% \label{fig:failure_case}
% \vspace{-1em}
% \end{figure}

\section{Conclusion}
In this paper, we introduce \textit{RealCraft}, a parameter-free zero-shot method for editing real videos based on the two key observations we made behind attention mechanisms. Our approach facilitates consistent shape-wise editing across up to 64 frames by simultaneously relaxing spatial-temporal attention and swapping cross-attention. Through quantitative and qualitative evaluation, we demonstrate \textit{RealCraft}'s effectiveness in consistent video editing, background transformation, precise shape editing, and pose preservation. A limitation of the proposed pipeline is its reliance on the performance of Stable Diffusion when extracting the cross-attention maps. Inaccurate maps may lead to sub-optimal edits. One example can be found in supplementary material. This problem can be alleviated by fine-tuning the pre-trained models. Given \textit{RealCraft}'s reliable performance in high-fidelity shape editing with enhanced temporal consistency, future developments could expand the guidance from prompt-only to multi-modal setting, thus expanding the control over object movement (e.g. trajectory, velocity) within scenes and offering users greater creative freedom in their edits.

\section{Acknowledgement}
This work was partially supported by the Wallenberg AI, Autonomous Systems and Software Program (WASP) funded by the Knut and Alice Wallenberg Foundation. The computations were enabled by the supercomputing resource Berzelius provided by the National Supercomputer Centre at Linköping University and the Knut and Alice Wallenberg Foundation, Sweden.

\clearpage

\bibliographystyle{plainnat}
\bibliography{main}
\end{document}